\documentclass{article}

\PassOptionsToPackage{authoryear,round}{natbib}
 \usepackage[preprint]{neurips_2025}

\usepackage[utf8]{inputenc} 
\usepackage[T1]{fontenc}    
\usepackage{hyperref}       
\usepackage{url}            
\usepackage{booktabs}       
\usepackage{amsfonts}       
\usepackage{nicefrac}       
\usepackage{microtype}      
\usepackage{xcolor}         
\usepackage{graphicx}
\usepackage{svg}
\usepackage{colortbl}
\usepackage{enumitem}

\hypersetup{
    colorlinks=true,   
    citecolor=neuripsaccentblue,    
    linkcolor=neuripsaccentblue,    
    urlcolor=neuripsaccentblue      
}
\title{OpenSeeker: Democratizing Frontier Search Agents by Fully Open-Sourcing Training Data}

\author{%
  \textbf{Yuwen Du\textsuperscript{1,*}}, \textbf{Rui Ye\textsuperscript{1,*,\#,\textdagger}},  \textbf{Shuo Tang\textsuperscript{1}}, \textbf{Xinyu Zhu\textsuperscript{1}}, \textbf{Yijun Lu\textsuperscript{1}}, \textbf{Yuzhu Cai\textsuperscript{1}}, \textbf{Siheng Chen\textsuperscript{1,\textdagger}} \\
  \textsuperscript{1}Shanghai Jiao Tong University, \textsuperscript{*}Equal Core Contributions, \textsuperscript{\#}Project Lead \\
  \textsuperscript{\textdagger}Corresponding Authors: yr991129@sjtu.edu.cn, sihengc@sjtu.edu.cn
}

\begin{document}

\maketitle

\begin{abstract}
  Deep search capabilities have become an indispensable competency for frontier Large Language Model (LLM) agents, yet the development of high-performance search agents remains dominated by industrial giants due to a lack of transparent, high-quality training data.
  This persistent data scarcity has fundamentally hindered the progress of the broader research community in developing and innovating within this domain.
  To bridge this gap, we introduce \textbf{OpenSeeker}, the first fully open-source search agent (i.e., model and data) that achieves frontier-level performance through two core technical innovations:
  (1) Fact-grounded scalable controllable QA synthesis, which reverse-engineers the web graph via topological expansion and entity obfuscation to generate complex, multi-hop reasoning tasks with controllable coverage and complexity.
  (2) Denoised trajectory synthesis, which employs a retrospective summarization mechanism to denoise the trajectory, therefore promoting the teacher LLMs to generate high-quality actions.
  Experimental results demonstrate that OpenSeeker, trained (a single training run) on only 11.7k synthesized samples, achieves state-of-the-art performance across multiple benchmarks including BrowseComp, BrowseComp-ZH, xbench-DeepSearch, and WideSearch.
  Notably, trained with simple SFT, OpenSeeker significantly outperforms the second-best fully open-source agent DeepDive (e.g., 29.5\% v.s. 15.3\% on BrowseComp), and even surpasses industrial competitors such as Tongyi DeepResearch (trained via extensive continual pre-training, SFT, and RL) on BrowseComp-ZH (48.4\% v.s. 46.7\%).
  We fully open-source the complete training dataset and the model weights to democratize frontier search agent research and foster a more transparent, collaborative ecosystem.

  \medskip
  \begin{flushleft}
    \begin{tabular}{@{}ll@{}}
      \includegraphics[width=1em]{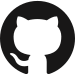} \quad \textbf{Code} & \href{https://github.com/rui-ye/OpenSeeker}{https://github.com/rui-ye/OpenSeeker} \\
      \includegraphics[width=1em]{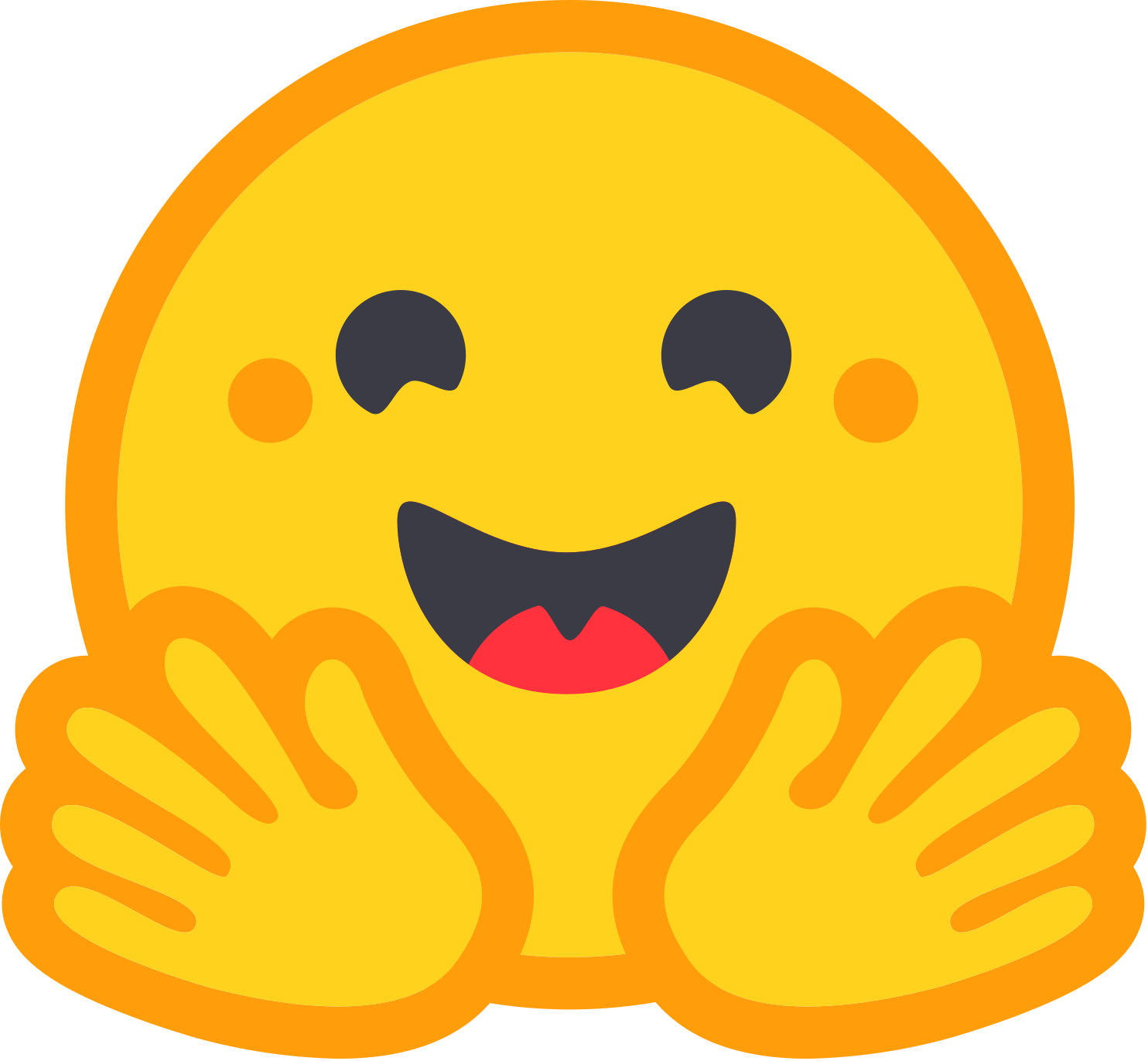} \quad \textbf{Data} & \href{https://huggingface.co/datasets/OpenSeeker/OpenSeeker-v1-Data}{https://huggingface.co/datasets/OpenSeeker/OpenSeeker-v1-Data} \\
      \includegraphics[width=1em]{figs/icons/huggingface.png} \quad \textbf{Model} & \href{https://huggingface.co/OpenSeeker/OpenSeeker-v1-30B-SFT}{https://huggingface.co/OpenSeeker/OpenSeeker-v1-30B-SFT} \\
    \end{tabular}
  \end{flushleft}
  \vspace{-4mm}
\end{abstract}

\begin{figure}[!h]
    \centering
    \vspace{-4mm}
    \includegraphics[width=1.0\linewidth]{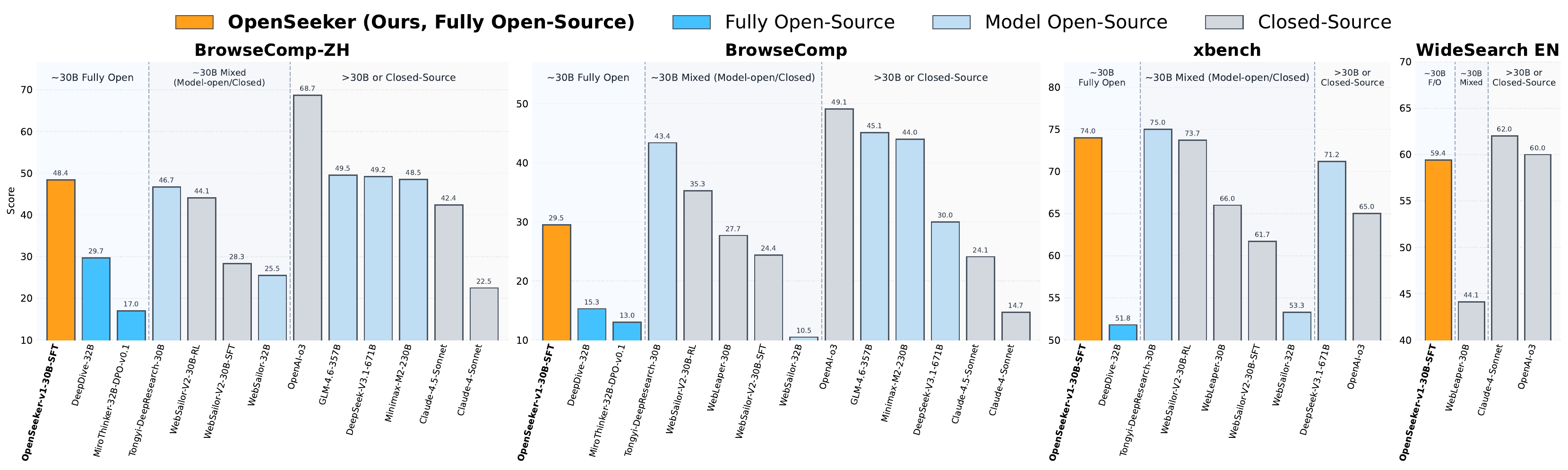}
    \vspace{-3mm}
    \caption{OpenSeeker stands out as the only fully open-source agent that achieves competitive performance on four search benchmarks, remarkably accomplishing this via simple SFT in a single training trial..}
    \label{fig:teaser}
\end{figure}

\section{Introduction}

In the era of information explosion, seeking accurate, real-time, and reliable information from the vast expanse of the internet has become a fundamental pillar of modern decision-making~\citep{marchionini1995information,given2023looking}.
Consequently, the ability to perform \textit{deep search} has emerged as a non-negotiable competency for frontier Large Language Model (LLM) agents~\citep{dr}.
The past year has witnessed a rapid rise in the development of search agents.
As recently as April 10, 2025, even the most advanced LLMs, such as OpenAI's o1~\citep{o1-preview}, struggled to surpass a score of 10 on the representative BrowseComp~\citep{bc_en} benchmark.
Yet, by March 2026, the landscape has shifted dramatically, with over ten agentic LLMs now exceeding the 50-point threshold~\citep{o3,team2026kimi,zeng2026glm}, signaling a new era of autonomous web intelligence.

However, despite this rapid progress, the training of high-performance search agents has remained a "closed-door game" played almost exclusively by well-funded corporate entities~\citep{gpt5.2,team2026kimi}.
The most capable search agents are currently dominated by proprietary models from giants such as Google and OpenAI.
While prominent labs including Kimi and Minimax have contributed open-weights models, they have remained silent regarding their training data.
Even within the research community, existing works either open-source the model without data~\citep{li2025websailorv2}, provide only a fraction of data~\citep{li2025websailor}, or fail to achieve competitive performance~\citep{lu2025deepdive}.
This persistent lack of complete high-quality training data has stifled the growth of the open-source community for nearly a year.

To bridge this gap, we, a purely academic team, introduce \textbf{OpenSeeker}, the first fully open-source search agent that achieves frontier-level performance in web search tasks.
OpenSeeker is not merely an open-weights model; it is a comprehensive democratization of the search agent pipeline, providing the community with all of training data, including both complex question-answer (QA) pairs and detailed trajectories.

The high-fidelity data behind OpenSeeker is powered by two core technical innovations: \textbf{fact-grounded scalable controllable QA synthesis} and \textbf{denoised trajectory synthesis}. 
Specifically, (1) our QA synthesis framework is designed to move beyond simple retrieval-based tasks that current models often solve through superficial pattern matching. 
To ensure queries demand genuine multi-hop reasoning, we reverse-engineer the web graph starting from randomly sampled seed pages within a massive web corpus. 
Specifically, we perform \textit{topological graph expansion} to identify interconnected information clusters, which are then distilled into \textit{entity subgraphs}. 
By applying \textit{entity obfuscation} to these subgraphs, we transform straightforward facts into complex reasoning puzzles that structurally mandate multi-step navigation. 
This approach ensures our data is \underline{fact-grounded} (anchored in real-world web topology), \underline{scalable} (terabytes of web archives available), and \underline{controllable} (modulating difficulty through subgraph complexity).
(2) Our trajectory synthesis method is designed to overcome the distractions inherent in raw web content.
During generation, we employ a secondary LLM to summarize preceding tool response, providing the teacher LLM with a cleaner/denoised history to produce superior reasoning and actions.
In the training phase, however, we supervise the model to predict these expert decisions while conditioning it on the \textit{original, raw} historical trajectory.
This decoupling compels the agent to internalize robust information-extraction capabilities, learning to ``see through the noise'' to identify the essential signals required for frontier-level performance.

To validate the efficacy of our data, we synthesize a dataset comprising 10.3k English and 1.4k Chinese samples and perform Supervised Fine-Tuning (SFT) on the Qwen3-30B-A3B~\citep{yang2025qwen3}.
Despite utilizing only SFT, OpenSeeker demonstrates remarkable competitiveness against models trained by corporate entities across benchmarks including BrowseComp~\citep{bc_en} (29.5\%), BrowseComp-ZH~\citep{bc_zh} (48.4\%), xbench-DeepSearch~\citep{xbench} (74.0\%), and WideSearch~\citep{wong2025widesearch} (59.4\% item F1)\footnote{It is worth highlighting that, due to resource constraints, these results are achieved in a single training run using default hyperparameters, without any heuristic filtering or hyperparameter optimization, leaving a large room for future research.}.
Notably, on the BrowseComp-ZH, OpenSeeker surpasses Alibaba’s Tongyi DeepResearch~\citep{team2025tongyi}, a model trained with extensive continual pre-training, SFT and RL (48.4 v.s. 46.7).
Among other models of equivalent scale trained via only SFT, our OpenSeeker achieves the \underline{best} performance on average, proving the high-quality nature of our training data.

Our primary contributions are summarized as follows:
\begin{itemize}[leftmargin=*]
    \item We propose two effective techniques: fact-grounded, scalable, controllable QA synthesis and denoised trajectory synthesis, enabling the automated generation of frontier-level training data.
    \item We develop and release OpenSeeker, a search agent that achieves state-of-the-art performance among open-source agents, matching or exceeding frontier solutions developed by corporate.
    \item We fully open-source the entire synthesis solution, the final training dataset (QA pairs and full trajectories), and the model weights, aiming to accelerating the development of search agents.
\end{itemize}

Ultimately, to the best of our knowledge, \textbf{OpenSeeker} represents the first work by a purely academic team to achieve state-of-the-art performance on frontier search benchmarks while fully open-sourcing the entirety of its training data.
Developed exclusively by an academic team, our work aims to democratize search intelligence by demonstrating that strategic data synthesis can effectively bridge the performance gap with industrial-scale efforts.
By providing full data transparency, we hope OpenSeeker serves as a catalyst for the research community to participate in a more open, collaborative, and healthy development of autonomous agents.

\section{Related Work}

The evolution of LLM-base search agents has shifted the paradigm of information retrieval from simple keyword matching to autonomous, multi-turn synthesis~\citep{marchionini1995information}.
Most contemporary search agents are architected upon the ReAct paradigm~\citep{yao2023react}, which utilizes a reasoning-action-observation loop to interact with web environments\footnote{While some parallel efforts focus on context management for agents~\citep{ye2025agentfold,team2025mirothinker}, our work primarily focuses on the fundamental challenge of data quality.}.
Historically, this path has been dominated by corporate entities.
(1) OpenAI's Deep Research~\citep{dr} pioneers the fully closed-source path, followed by a series of proprietary agents including Kimi-Researcher~\citep{kimiresearcher}, Gemini’s Deep Research~\citep{gemini2.5}, and Perplexity’s Deep Research~\citep{perplexity_dr}.
(2) Within the past six months, a wave of "open-weights" models capable of search has emerged, such as the Kimi K2/2.5 series~\citep{kimi-k2,team2026kimi}, Zhipu GLM 4.5-5~\citep{zeng2025glm,zeng2026glm}, MiniMax M2-2.5~\citep{minimax2,minimax2.5}, and Alibaba’s Tongyi DeepResearch~\citep{team2025tongyi}.
However, none of these industrial efforts have disclosed their training data, effectively maintaining a "data moat" that preserves frontier performance as a corporate secret.
(3) While the research community has made significant strides with frameworks such as WebDancer~\citep{wu2025webdancer}, WebSailor~\citep{li2025websailor}, WebSailor-V2~\citep{li2025websailor}, WebLeaper~\citep{tao2025webleaper}, AgentFounder~\citep{su2026scaling}, DeepDive~\citep{lu2025deepdive}, and MiroThinker~\citep{2025mirothinker}, they either lack public releases, provide only a small fraction of the data, or suffer from low data fidelity that fails to achieve competitive performance.

This status quo has left the research community lacking of the high-quality data necessary to train high-performance agents.
\textbf{OpenSeeker} explicitly addresses this void by fully open-sourcing its entire synthesis pipeline and high-fidelity training data, democratizing the "recipe" for frontier search intelligence\footnote{We discuss with two concurrent works in Section~\ref{app:concurrent_work}.}.
\textbf{To the best of our knowledge, OpenSeeker represents the first work by a purely academic team to achieve state-of-the-art performance on frontier search benchmarks while simultaneously open-sourcing the full training data.}
Notably, our SOTA results are achieved within a single training trial without any iterative refinement, underscoring the high quality of our synthesized data and leaving substantial room for future exploration.

\section{Methodology}

\subsection{Overview \& Problem Formulation}
Our primary objective is to synthesize a high-fidelity dataset $\mathcal{D} = \{(q, y, \tau^*)\}$ comprising complex queries $q$, ground truth answers $y$, and optimal tool-use trajectories $\tau^*$. This dataset aims to empower an agent $\pi_\theta$ to master long-horizon tool invocation for deep search tasks.

We model the web as a directed graph $\mathcal{G} = (\mathcal{V}, \mathcal{E})$, where $\mathcal{V}$ denotes web pages and $\mathcal{E}$ denotes hyperlinks. The synthesis challenge is to derive pairs $(q, y)$ from $\mathcal{G}$ such that solving $q$ necessitates a trajectory $\tau = [a_1, o_1, \dots, a_T, o_T]$ of length $T \gg 1$, where $a_t$ are search actions and $o_t$ are observations. We argue that to effectively train deep search agents, one must address two pivotal challenges: (1) \textbf{High-difficulty QA:} Only sufficiently complex queries compel the system to engage in a rigorous multi-turn interaction cycle involving ``Reasoning $\rightarrow$ Tool Call $\rightarrow$ Tool Response''. This process is essential to generate long-horizon trajectories characterized by explicit decision points and extended tool invocation chains. (2) \textbf{High-quality trajectories:} The synthesis of solution paths must rely on stable and reproducible methods to ensure that the distilled training signals represent ``correct and generalizable'' strategies rather than accidental successes derived from stochastic sampling.

To address these, we propose a \textbf{fact-grounded
scalable controllable QA synthesis} framework and a \textbf{denoised trajectory synthesis} method. The QA synthesis framework operates on the premise of \textit{reverse-engineering} the reasoning graph: we first identify a latent inference path within $\mathcal{G}$ and then construct a question $q$ that structurally mandates traversing this path. Complementarily, our trajectory synthesis method utilizes dynamic context denoising to generate clear reasoning and precise tool calls. By subsequently training on raw trajectories, we enable the agent to intrinsically learn to denoise and extract relevant information from noisy tool responses.

\subsection{Fact-Grounded
Scalable Controllable QA Synthesis}
\begin{figure*}[t]
\centering
\includegraphics[width=1\linewidth]{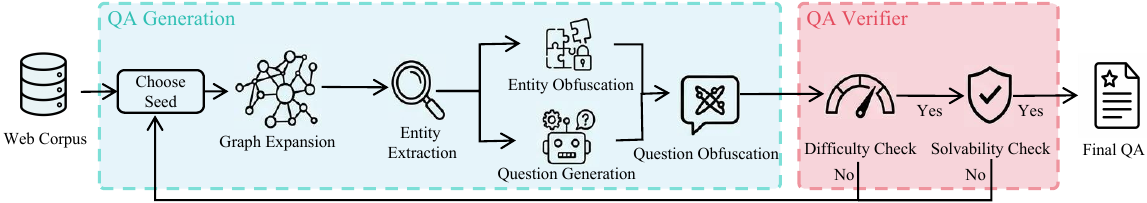}
   \caption{Overview of Fact-grounded scalable controllable QA synthesis. The pipeline begins with \textit{Graph Expansion}, where a seed node is expanded into a subgraph of connected pages. \textit{Entity Extraction} then distills key information themes into a structured Entity Subgraph. A generator synthesizes complex initial questions conditioned on this structure (\textit{Question Generation}), ensuring multi-hop reasoning requirements. To enhance difficulty, we apply \textit{Entity Obfuscation} to vagueify specific terms, finally producing a challenging question that necessitates deep graph traversal to solve. }
   \label{fig:QA}
   \vspace{-0.3cm}
\end{figure*}

We engineer a pipeline to construct question-answer pairs $(q, y)$ directly from the web graph $\mathcal{G}$, as shown in Figure \ref{fig:QA}. By leveraging intrinsic connectivity, we transform static hyperlinks into dynamic reasoning paths, ensuring factual grounding and controllable complexity. This scalable framework operates in two distinct phases: \textit{Generative Construction} to synthesize candidate pairs, and \textit{Dual-Criteria Verification} to rigorously filter for difficulty and solvability.

\subsubsection{Generative Construction: From Graph to Question}

\textbf{Graph Expansion.} To mimic the natural process of information discovery where one clue leads to another, we initiate the pipeline by sampling a seed node $v_{seed} \sim \mathcal{V}$. Recognising that complex questions rarely reside on a single isolated page, we expand from $v_{seed}$ by traversing its outgoing edges in $\mathcal{E}$ to gather a set of $k$ connected nodes. This forms a local dependency subgraph $\mathcal{G}_{sub} = \{v_{seed}\} \cup \{v_i | (v_{seed}, v_i) \in \mathcal{E}\}_k$, which serves as a coherent, topologically-linked knowledge base for problem construction.

\textbf{Entity Extraction.} Synthesizing complex questions necessitates utilizing a generative model to reference the information cluster within the expanded subgraph $\mathcal{G}_{sub}$. However, the raw content of these nodes often contains excessive noise that can distract the generation model. To sharpen the focus, we identify the central theme $y_{theme}$ of $v_{seed}$ and execute an extraction function. This function distills a set of key entities from across the subgraph that are directly or indirectly related to the central theme $y_{theme}$, and reorganizes them into a condensed \textit{Entity Subgraph} $\mathcal{G}_{entity}$. In this graph, nodes represent the extracted entities and edges preserve the original topological connections efficiently. This step effectively abstracts $\mathcal{G}_{sub}$ into a dense relational structure, removing textual noise while retaining the essential logic paths.

\textbf{Question Generation.} To prevent the generation of questions that can be solved by simple look-up, we employ a generator $P_{gen}$ to synthesize an initial question $q_{init}$ conditioned explicitly on the structure of the Entity Subgraph $\mathcal{G}_{entity}$. We impose a hard structural constraint: the derivation of $y_{theme}$ from $q_{init}$ must necessitate traversing multiple edges within $\mathcal{G}_{entity}$. This explicitly forces the agent to engage in sequential multi-node deductive reasoning rather than single-step retrieval.

\textbf{Entity Obfuscation.} The synthesized questions are intended to drive agents to perform multi-step ReAct reasoning. However, agents often exploit specific keywords to shortcut the reasoning process via direct search. To simulate realistic user ambiguity and dismantle these shortcuts, we apply an obfuscation operator $\Phi$ directly to the entity nodes in $\mathcal{G}_{entity}$. Concrete entities $e$ are mapped to vague, descriptive references $\tilde{e} = \Phi(e)$. This transformation yields a \textit{Fuzzy Entity Subgraph} $\tilde{\mathcal{G}}_{entity}$, where the structural connectivity remains intact but the semantic nodes now demand disambiguation.

\textbf{Question Obfuscation.} The pipeline culminates in generating the final question $\tilde{q}$ by taking the initial question $q_{init}$ and the fuzzy entity subgraph $\tilde{\mathcal{G}}_{entity}$ as inputs. This separation allows the generator to reference the pre-obfuscated descriptions in $\tilde{\mathcal{G}}_{entity}$ directly, thereby focusing exclusively on synthesizing the complex question structure. The generator rewrites $q_{init}$ to incorporate the ambiguous descriptions while preserving the original reasoning logic, with the target answer remaining the invariant $y = y_{theme}$.

\subsubsection{Dual-Criteria Verification via Rejection Sampling}
To ensure the synthesized pair $(\tilde{q}, y)$ is both challenging and valid, we employ a rejection sampling scheme based on two indicator functions:

\textbf{(1) Criterion 1: difficulty (strict tool necessity).} Let $\pi_{base}$ be a strong foundation model. We define the difficulty condition as $\mathbb{I}[\pi_{base}(\tilde{q}) \neq y]$, where $\pi_{base}$ generates an answer in a closed-book setting (no external tools). If the model answers correctly using only parametric memory, the question is discarded. This guarantees that $\tilde{q}$ necessitates external information seeking.

\textbf{(2) Criterion 2: solvability (logical consistency).} We define the solvability condition as $\mathbb{I}[\pi_{base}(\tilde{q} | \mathcal{G}_{entity}) = y]$. Here, the model is provided with the full content of the Entity Subgraph $\mathcal{G}_{entity}$ as context (oracle setting). If the model fails to derive $y$, it implies the reasoning path is broken or hallucinated. Such samples are rejected to strictly enforce logical validity.

\subsubsection{Discussions}

Our data synthesis paradigm fundamentally advances agent training through three core strengths:

(1) \textbf{Factual grounding:}
By anchoring queries in the real web’s topology rather than relying on LLM generation, hallucination risks are significantly mitigated, if not entirely eliminated.
Every training example is strictly grounded to verifiable, real-world data. 

(2) \textbf{Scalability:}
In this work, we leverage $\sim$68GB English and $\sim$9GB Chinese web data to testify our solution, demonstrating that it suffices to synthesize high-quality QA pairs for training high-performance search agents.
With TB-scale web archives still largely untapped, our pipeline transforms the open web into an inexhaustible source.
By continuously varying seed pages or adjusting graph configurations, we can generate an (almost) infinite stream of diverse, non-repeating samples, ensuring no data bottlenecks for model scaling. 

(3) \textbf{Controllability:}
In our solution, task difficulty is a deliberate design choice rather than a random variable
By tuning the subgraph size (k), we can calibrate reasoning complexity and information coverage.
This enables us to build tailored curriculums that progressively guide agents from straightforward retrieval to sophisticated, multi-hop investigations.

\subsection{Denoised Trajectory Synthesis}

Constructing high-quality search trajectories requires strictly balancing information retention with context window constraints. In web-scale search, raw observations are often dominated by irrelevant noise. To address this, we propose a synthesis framework that technically decouples the \textit{generation context} (Teacher) from the \textit{training context} (Student), employing a dynamic context denoising strategy.

\begin{figure*}[t]
\centering
\includegraphics[width=1\linewidth]{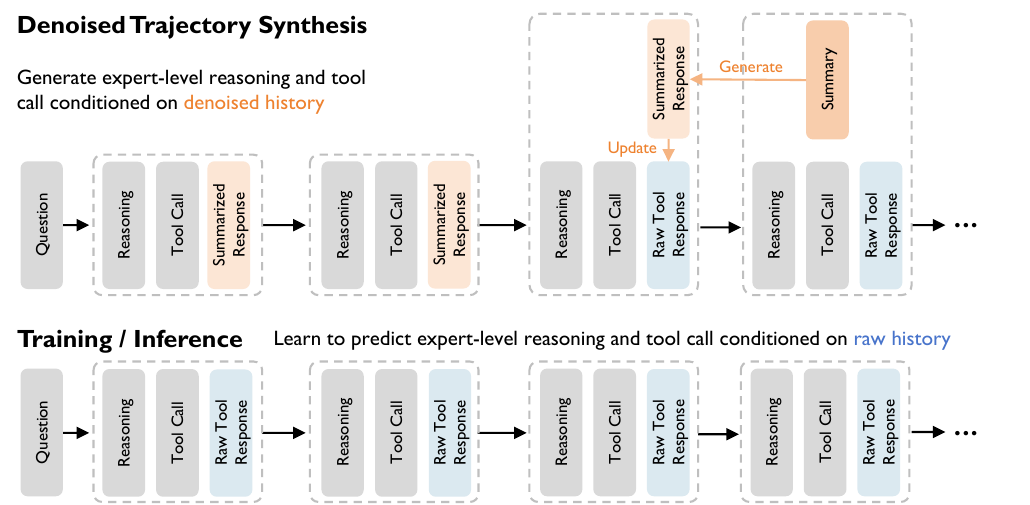}
   \caption{Overview of Denoised Trajectory Synthesis. We employ a retrospective summarization mechanism where, after each tool call, the raw tool response from the previous turn is condensed into a `Summarized Response' that replaces the original raw tool response in the history window. This cleaner context enables the teacher to generate high-quality reasoning and actions. Note the asymmetry: while synthesis relies on summarized context, the training and inference phases operate on raw tool response to force the model to learn intrinsic denoising capabilities.}
   \label{fig:traj}
\end{figure*}

\subsubsection{Problem Formulation}
Let a search trajectory be defined as a sequence $\tau = [q, (r_1, a_1, o_1), \dots, (r_T, a_T, o_T), y]$, where $q$ is the question, $r_t$ is the reasoning step (chain-of-thought), $a_t$ is the action (tool call), and $o_t$ is the observation (tool response) at turn $t$, culminating in the final answer $y$. Our goal is to synthesize specific reasoning paths $r_t$ and actions $a_t$ that optimally lead to $y$.

\subsubsection{Synthesis via Dynamic Context Denoising}
During trajectory synthesis, we employ a \textbf{retrospective summarization} mechanism. This ensures that the agent utilizes the complete information from the immediate past while maintaining a concise long-term memory. Formally, at turn $t$, the agent generates the reasoning and action pair $(r_t, a_t)$ based on the current context $\mathcal{H}_t$. Our context construction follows a ``Summarized History + Raw Recent'' protocol:
\begin{equation}
    \mathcal{H}_t = \{q, \underbrace{(r_1, a_1, s_1), \dots, (r_{t-2}, a_{t-2}, s_{t-2})}_{\text{Summarized Long-Term History}}, \underbrace{(r_{t-1}, a_{t-1}, o_{t-1})}_{\text{Raw Short-Term Context}}\}
\end{equation}
where $s_i = \text{Summarize}(o_i | \text{context})$ represents the compressed semantic summary of the observation $o_i$. This mechanism operates in a two-phase cycle:

(1) \textbf{Decision phase (information usage):} To generate the current decision $(r_t, a_t)$, the agent is provided with $\mathcal{H}_t$, which includes the \textit{full raw observation} $o_{t-1}$ from the immediately preceding step. This guarantees that the agent has access to all potential signals in the most recent observation to inform its next move, preventing premature information loss.

(2) \textbf{Compression phase (context denoising):} Once step $t$ is concluded and a new observation $o_t$ is obtained, the system retrospectively invokes a summarizer to compress the previous observation $o_{t-1}$ into $s_{t-1}$. This summary $s_{t-1}$ then replaces $o_{t-1}$ in the long-term history for the next step $\mathcal{H}_{t+1}$. This rolling window approach effectively filters noise and denoises the context, enabling the generation of extremely long horizons without performance degradation.

\subsubsection{Asymmetric Context Training for Robust Denoising}
To cultivate robustness in the final agent, we define a strategic asymmetry between the data format used for synthesis and that used for training, as shown in Figure \ref{fig:traj}. (1) \textbf{Synthesis data (teacher):} The trajectories are generated using the clean, denoised context $\mathcal{H}_t$ containing summaries. This acts as a scaffold, allowing the teacher model to produce ``golden'' reasoning paths unencumbered by excessive noise. (2) \textbf{Training data (student):} For the final training dataset, we strip away the summaries and revert to the full raw context:
    \begin{equation}
        \mathcal{H}^{train}_t = \{q, (r_1, a_1, o_1), \dots, (r_{t-1}, a_{t-1}, o_{t-1})\}
    \end{equation}
The student model is supervised to predict the optimal $r_t, a_t$ (derived from the Teacher) given the noisy, raw context $\mathcal{H}^{train}_t$. This forces the student to implicitly learn the \textit{denoising} and \textit{information extraction} capabilities, effectively internalizing the context denoising logic within its own parameters to handle real-world unstructured data.

\section{Experiments}
\label{sec:experiments}


\subsection{Experimental Setup}

\textbf{Implementation.} We develop OpenSeeker, a deep search agent initialized from Qwen3-30B-A3B-Thinking-2507~\citep{qwen3thinking}, featuring 30B total parameters with 3B activated during prediction. The maximum tool call limit is set to 200, with any trajectory exceeding this threshold being forcibly terminated. The context window size is set to 256k. 
Each training sample comprises a user question $q$ and a sequence of raw reasoning steps, tool calls, and \textit{full, uncompressed} tool responses.
Note that due to resource constraints, we only train the model for a single run, without any heuristic data filtering or hyperparameter-tuning for training, leaving a large improving room for future research.

\textbf{Benchmarks.} We evaluate OpenSeeker on four key benchmarks: BrowseComp~\citep{bc_en} and BrowseComp-ZH~\citep{bc_zh}, which test multi-step navigation and hard information location in English and Chinese, respectively (we evaluate BrowseComp results on a subset of 200 samples due to resource constraints); xbench-DeepSearch~\citep{xbench}, assessing complex deep research capabilities like planning and synthesis; and WideSearch~\citep{wong2025widesearch}, measuring reliability in broad information seeking across extensive sources.

\textbf{Baselines.} To assess the efficacy of OpenSeeker, we compare it against a broad spectrum of state-of-the-art systems categorized into three groups: (1) closed-source proprietary models, representing the industry upper bound (e.g., Claude series~\citep{claude4}, OpenAI-o3~\citep{o3}, OpenAI Deep Research~\citep{dr}, GPT-5-High~\citep{Singh2025GPT5SystemCard}); (2) large-scale open-source models, comprising massive parameter systems such as Kimi-K2~\citep{kimi-k2}, DeepSeek (V3.1/V3.2~\citep{DeepSeek2025V32}), GLM-4 (4.6/4.7)~\citep{5team2025glm45agenticreasoningcoding}, Minimax-M2~\citep{MiniMaxM2}, and LongCat-Flash~\citep{longcat2025flash}; and (3) $\sim$30B models, which serve as direct, comparable-scale benchmarks. This group includes representative search agents such as MiroThinker series~\citep{2025mirothinker}, DeepDive-32B~\citep{lu2025deepdive}, WebDancer~\citep{wu2025webdancer}, WebSailor-V2~\citep{li2025websailorv2}, WebLeaper~\citep{tao2025webleaper}, and Tongyi DeepResearch~\citep{Li2025TongyiDeepResearch}. Baseline performance metrics are derived from their respective technical reports or public leaderboards.

\begin{table*}[t]
    \centering
    \small
        \caption{Comparisons among our OpenSeeker and other search agents. `\# Samples' denotes the number of total training data samples; `\# OS Samples' denotes the number of open-source data samples; `Training' denotes training techniques (CPT: continual pre-training, SFT: supervised fine-tuning, RL: reinforcement learning); `Academic' denotes whether conducted by pure academic team ($\checkmark$: Yes, $\times$: No); `BC-ZH' denotes BrowseComp-ZH; `WideSearch' denotes item F1 result on the English subset. With simple SFT, OpenSeeker-v1-30B-SFT even surpasses Tongyi DeepResearch on BrowseComp-ZH which is trained via CPT, SFT, and RL. \textbf{Among all SFT-based agents, OpenSeeker performs significantly best.}}
        \label{tab:deep_research_complete}
        \resizebox{\textwidth}{!}{
        \begin{tabular}{lcccccccc}
            \toprule
            \textbf{Model Name} & \textbf{\# Samples} & \textbf{\# OS Samples} & \textbf{Training} & \textbf{Academic} & \textbf{BrowseComp} & \textbf{BC-ZH} & \textbf{xbench} & \textbf{WideSearch} \\
            \midrule
            
            \rowcolor{blue!8}\multicolumn{9}{c}{\emph{\textbf{Closed-Source Proprietary Models}}} \\
            Claude-4-Sonnet & ? & 0 & ? & $\times$ & 14.7 & 22.5 & - & 62.0 \\
            Claude-4.5-Sonnet & ? & 0 & ? & $\times$ & 24.1 & 42.4 & - & - \\
            Claude-4-Opus & ? & 0 & ? & $\times$ & 18.8 & 37.4 & - & - \\
            OpenAI-o3 & ? & 0 & ? & $\times$ & 49.1 & 68.7 & - & 60.0 \\
            OpenAI Deep Research & ? & 0 & ? & $\times$ & 51.5 & 42.9 & - & - \\
            GPT-5-High & ? & 0 & ? & $\times$ & 54.9 & 63.0 & - & - \\
            \midrule

           \rowcolor{blue!8}\multicolumn{9}{c}{\emph{\textbf{Open-Source Models > 30B}}} \\
            Kimi-K2-Instruct-1T  & ? & 0 & ? & $\times$ & 14.1 & 28.8 & - & 59.9 \\
            DeepSeek-V3.1-671B & ? & 0 & ? & $\times$ & 30.0 & 49.2 & 71.2 & - \\
            DeepSeek-V3.2-671B & ? & 0 & ? & $\times$ & 51.4 & 65.0 & - & - \\
            GLM-4.6-357B & ? & 0 & ? & $\times$ & 45.1 & 49.5 & - & - \\
            GLM-4.7-357B & ? & 0 & ? & $\times$ & 52.0 & 66.6 & - & - \\
            Minimax-M2-230B & ? & 0 & ? & $\times$ & 44.0 & 48.5 & - & - \\
            \midrule

            \rowcolor{blue!8}\multicolumn{9}{c}{\emph{\textbf{$\sim$30B Models}}} \\
            WebDancer-32B & ? & 0 & SFT + RL & $\times$ & 3.8 & 18.0 & - & - \\
            MiroThinker-32B-v0.1 & 147 k & 147 k & SFT & $\times$ & 10.6 & 13.8 & - & - \\
            MiroThinker-32B-v0.1 & 147 k & 147 k & SFT + RL & $\times$ & 13.0 & 17.0 & - & - \\
            DeepDive-32B & 4.1 k & 4.1 k & SFT + RL & $\times$ & 15.3 & 29.7 & 51.8 & - \\
            WebSailor-32B & ? & 0 & SFT + RL & $\times$ & 10.5 & 25.5 & 53.3 & - \\
            WebSailor-V2-30B & ? & 0 & SFT & $\times$ & 24.4 & 28.3 & 61.7 & - \\
            WebSailor-V2-30B & ? & 0 & SFT + RL & $\times$ & 35.3 & 44.1 & 73.7 & - \\
            WebLeaper-30B & 15 k & 0 & SFT & $\times$ &27.7 & - & 66.0 & 44.1 \\
            Tongyi DeepResearch & ? & 0 & CPT + SFT + RL & $\times$ & 43.4 & 46.7 & 75.0 & - \\
            \cmidrule(lr){1-9}
            
            \textbf{OpenSeeker-v1-30B-SFT} & 11.7 k & 11.7 k & SFT & \checkmark & 29.5 & 48.4 & 74.0 & 59.4 \\
            \bottomrule
        \end{tabular}
        }
\end{table*}

\subsection{Main Results}

\textbf{Outperforming resource-intensive industry baselines.} As shown in Table~\ref{tab:deep_research_complete}, our primary evaluation compares OpenSeeker against a spectrum of proprietary and open-source models, highlighting its superior efficiency and competitiveness achieved through just a single training run.
Despite utilizing a modest dataset of only 11.7k samples and a standard SFT protocol, OpenSeeker consistently rivals or exceeds the performance of models backed by massive corporate resources. A standout result is observed on the BrowseComp-ZH benchmark, where OpenSeeker achieves a score of 48.4, surpassing Alibaba's Tongyi DeepResearch (46.7). This is particularly significant given that Tongyi DeepResearch employs a complex, resource-heavy training pipeline involving Continual Pre-Training (CPT), SFT, and Reinforcement Learning (RL), whereas OpenSeeker relies solely on high-quality SFT data. 

\textbf{Superior performance under identical training setup.} As shown in Table~\ref{tab:model_comp_sft}, when evaluated under the same SFT training protocol in the $\sim$30B parameter class, OpenSeeker demonstrates a decisive advantage, highlighting the effectiveness of our data synthesis method. Most notably on BrowseComp-ZH, OpenSeeker (48.4) outperforms the runner-up WebSailor-V2-SFT (28.3) by nearly 20\%. Additionally, models such as MiroThinker-32B-v0.1-SFT (13.8) lag significantly behind, confirming that data \textit{quantity} (e.g., 147k for MiroThinker) is secondary to data \textit{quality}. Our \textit{Denoised Trajectory Synthesis} effectively teaches the model to denoise and extract ``needle-in-a-haystack'' information from raw, noisy web observation, a capability that standard SFT datasets often fail to cultivate.

\begin{table}[t]
    \centering
    \small 
    \caption{Performance comparison of different models trained via SFT. Our OpenSeeker consistently and significantly performs the best across four benchmarks with only 11.7k training samples.}
    \label{tab:model_comp_sft}
    \scalebox{0.85}{ 
    \begin{tabular}{l c c c c c c c}
        \toprule
        \textbf{Data} & \textbf{\# Samples} & \textbf{\# OS Samples} & \textbf{Academic} & \textbf{BrowseComp} & \textbf{BC-ZH} &\textbf{xbench} & \textbf{WideSearch-EN} \\
        \midrule
        DeepDive-32B & 4.1 k & 4.1 k & $\times$ & 9.5 & 23.0 & 48.5 & - \\
        MiroThinker-32B-v0.1 & 147 k & 147 k & $\times$ & 10.6 & 13.8 & - & - \\
        WebSailor-V2-30B & ? & 0 & $\times$ & 24.4 & 28.3 & 61.7 & - \\
        WebLeaper-30B & 15 k & 0 & $\times$ &27.7 & - & 66.0 & 44.1 \\
        \midrule
        \textbf{OpenSeeker-v1-30B-SFT} & 11.7 k & 11.7k &  \checkmark & \textbf{29.5} & \textbf{48.4} &\textbf{74.0} & \textbf{59.4} \\
        \bottomrule
    \end{tabular}
    }
\end{table}

\textbf{Superior performance with comparable data volume.} To further isolate the contribution of our synthesis methodology, we compare OpenSeeker against various configurations of WebSailor-V2 and WebLeaper data. As shown in Table~\ref{tab:model_comp_small}, despite using a comparable or even smaller volume of data ($\sim$11.7k samples vs. 10k--15k samples), OpenSeeker demonstrates superior performance across all benchmarks. Specifically, on xbench and WideSearch, it outperforms the best baseline combination (utilizing 15k samples) by nearly 8\% (74.0) and 15\% (59.4), respectively. This result strongly validates the high quality and efficiency of our data, demonstrating that our synthesized samples provide significantly more effective supervision signals. In stark contrast to baselines relying on proprietary, company-synthesized datasets, OpenSeeker, developed by a purely academic research team, achieves this efficiency by leveraging our independently synthesized high-difficulty QA and high-quality denoised trajectories, while fully open-sourcing the entire dataset to the community.


\begin{table}[t]
    \centering
    \small 
    \caption{Performance comparison under comparable data volumes. OpenSeeker achieves significant advantages across three benchmarks, demonstrating the high quality of our data.}
    \label{tab:model_comp_small}
    \scalebox{0.8}{ 
    \begin{tabular}{l c c c c c c}
        \toprule
        \textbf{Data} & \textbf{\# Samples} & \textbf{\# OS Samples} & \textbf{Developer} & \textbf{BrowseComp} & \textbf{xbench} & \textbf{WideSearch-EN} \\
        \midrule
        WebSailor-V2-10k & 10k & 0 &  Tongyi & 24.50 & 62.67 & 38.91 \\
        WebSailor-V2-5k + WebLeaper-Basic-5k & 10k & 0 &  Tongyi & 20.67 & 58.33 & 32.26 \\
        WebSailor-V2-5k + WebLeaper-Union-5k & 10k & 0 &  Tongyi & 27.50 & 62.33 & 41.70 \\
        WebSailor-V2-5k + WebLeaper-Reverse-Union-10k & 15k & 0 &  Tongyi & 27.67 & 66.00 & 44.07 \\
        \midrule
        \textbf{OpenSeeker-v1-Data-11.7k} & 11.7 k & 11.7k &  Academic & \textbf{29.50} & \textbf{74.00} & \textbf{59.40} \\
        \bottomrule
    \end{tabular}
    }
\end{table}

\begin{figure}[t]
    \centering
    \includegraphics[width=\linewidth]{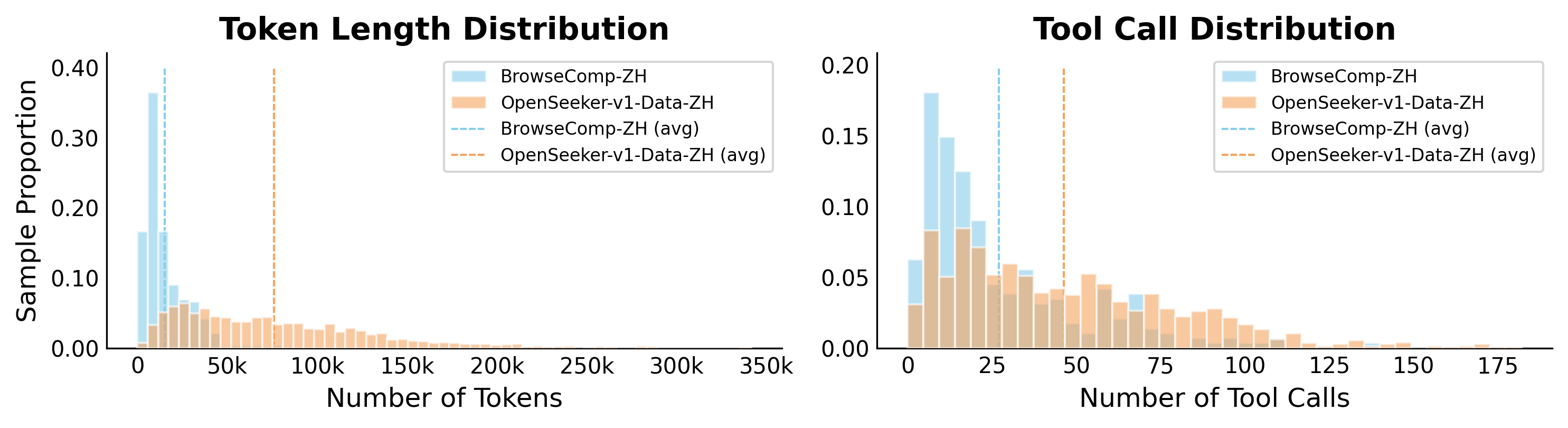}
    \caption{Comparison of difficulty between OpenSeeker-v1-Data-ZH and BrowseComp-ZH using the same model for inference. OpenSeeker-v1-Data-ZH exhibits significantly higher average token counts and tool call counts than BrowseComp-ZH.}
    \label{fig:data_stats_zh}
\end{figure}

\begin{figure}[t]
    \centering
    \includegraphics[width=\linewidth]{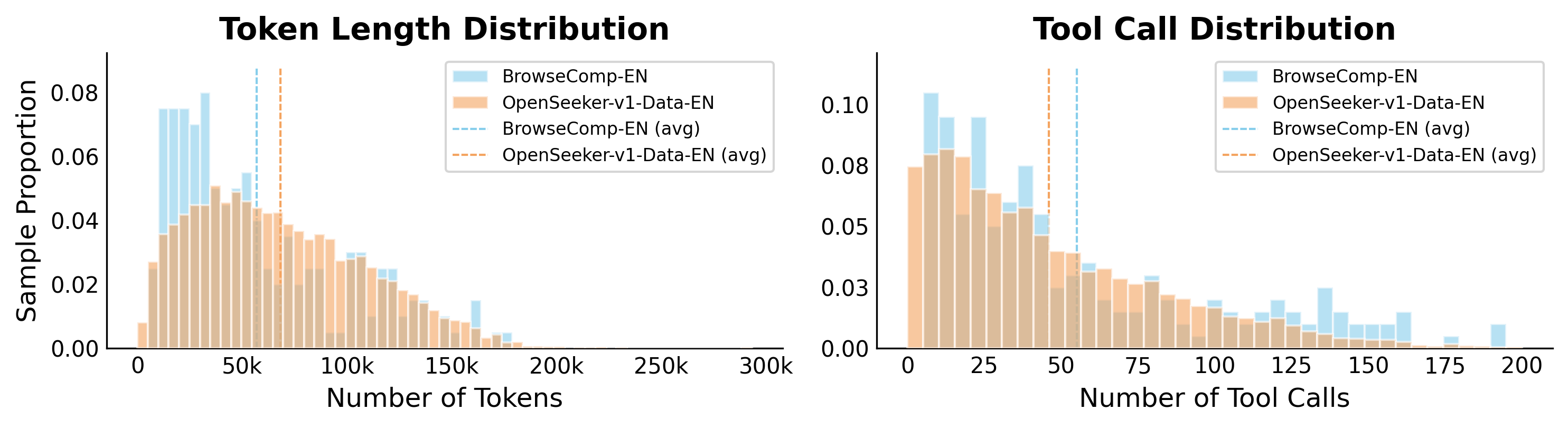}
    \caption{Comparison of difficulty between OpenSeeker-v1-Data-EN and BrowseComp-EN using the same model for inference. OpenSeeker-v1-Data-EN exhibits difficulty comparable to that of BrowseComp-EN.}
    \label{fig:data_stats_en}
\end{figure}

\textbf{Data statistics analysis.} To quantitatively contrast the difficulty of our synthesized data with that of standard benchmarks, we employ an open-source model to perform inference on both our synthesized samples and the BrowseComp benchmarks (including BrowseComp-ZH and BrowseComp-EN). The comparison reveals that our synthesized data matches or even exceeds the difficulty of established benchmarks.
Notably, although our Chinese dataset contains only approximately 1.4k samples, its complexity significantly surpasses that of BrowseComp-ZH.
As illustrated in Figure~\ref{fig:data_stats_zh}, our synthesized Chinese data averages \textbf{46.35} tool calls per trajectory with an average token length of \textbf{76.1k}, whereas BrowseComp-ZH averages only \textbf{26.98} tool calls and \textbf{15.1k} tokens.
This not only demonstrates that our problems are inherently more challenging but also validates that despite the limited data volume, its high fidelity and complexity directly contribute to superior performance on Chinese benchmarks.
Due to resource constraints, our English data has not yet been updated to the latest QA standards, resulting in slightly lower difficulty compared to the Chinese data (see Figure~\ref{fig:data_stats_en}).
We expect to release an updated version in the near future.

\section{Discussions}
\textbf{Breaking the corporate data monopoly.}
For a long period, the development of high-performance search agents has been a ``closed-door game'' dominated by tech corporate, with high-quality data serving as their primary moat.
Concurrently, existing open-source datasets often suffer from poor quality and inadequate reasoning complexity, leaving the academic community ill-equipped to train truly capable, frontier-level search models.
OpenSeeker addresses this critical bottleneck.
By open-sourcing a high-fidelity dataset that enables frontier-level performance, we provide the community with the necessary resources to replicate and build upon industrial-grade capabilities, breaking the long-standing ``data moat''.

\textbf{Future work.}
While our current work demonstrates significant effectiveness, it represents merely a lower bound of OpenSeeker's potential.
Due to resource constraints, we can only train for a single run, limiting not only the verification of effectiveness on more challenging data, but also the exploration of various parameters and data filtering strategies. 
In the next phase, we aim to optimize data distributions, implement rigorous quality filtering, and generate training data of even higher complexity to push the boundaries of performance. 
Furthermore, we plan to extend the agent's capabilities beyond pure web search by integrating a more diverse set of tools and data sources, ultimately advancing toward a more versatile and generalist agentic framework. 

\section{Conclusions}
The growth of the open-source search agent community has long been stifled by the monopoly of high-quality training data held by industrial corporations.
To bridge this gap, \textbf{OpenSeeker} represents the first work by a purely academic team to achieve state-of-the-art performance on frontier search benchmarks while simultaneously open-sourcing the full training data. 
Notably, our SOTA results are achieved using only 11.7k synthesized samples through a single supervised fine-tuning run, surpassing industrial baselines that rely on extensive resources and complex training pipelines.
This efficiency validates the effectiveness of our proposed \textit{fact-grounded scalable controllable QA synthesis} and \textit{denoised trajectory synthesis} methods in producing high-fidelity training data.
By openly releasing our the complete dataset, and model weights, we aim to dismantle the data barriers in this domain and foster a more inclusive, transparent, and collaborative ecosystem for future search agent research.

\medskip
{
\bibliographystyle{plainnat}
\bibliography{ref}
}

\clearpage
\appendix
\section{Concurrent Works}
\label{app:concurrent_work}

As the field moves toward more transparent search agent development, several concurrent efforts have emerged, yet they differ significantly from OpenSeeker in methodology and openness. 
(1) OpenResearcher~\citep{li2025openresearcher} primarily aggregates QA pairs from existing open-source datasets and constructs trajectories within simulated environments. 
In contrast, OpenSeeker creates entirely new, high-difficulty QA pairs via our graph-grounded synthesis and collects trajectories within \textit{real-world} web environments to ensure better generalizability. 
Furthermore, OpenSeeker demonstrates superior data quality, outperforming OpenResearcher (that is trained using 96k samples) with only 11.7k high-fidelity samples. 
(2) RedResearcher~\citep{chu2026redsearcher} adopts a multi-stage pipeline involving mid-training, SFT, and Reinforcement Learning (RL). 
However, it lacks full transparency regarding its training protocol and only provides a partial release of its SFT and RL data. 
Crucially, both OpenResearcher and RedResearcher involve significant corporate participation. 
OpenSeeker distinguishes itself as the first academic-led initiative to achieve state-of-the-art performance with a lean, SFT-only approach and 100\% data transparency, proving that strategic data synthesis can bridge the gap traditionally filled by massive corporate compute and iterative RL cycles.

\begin{figure}[t]
    \centering
    \includegraphics[width=\linewidth]{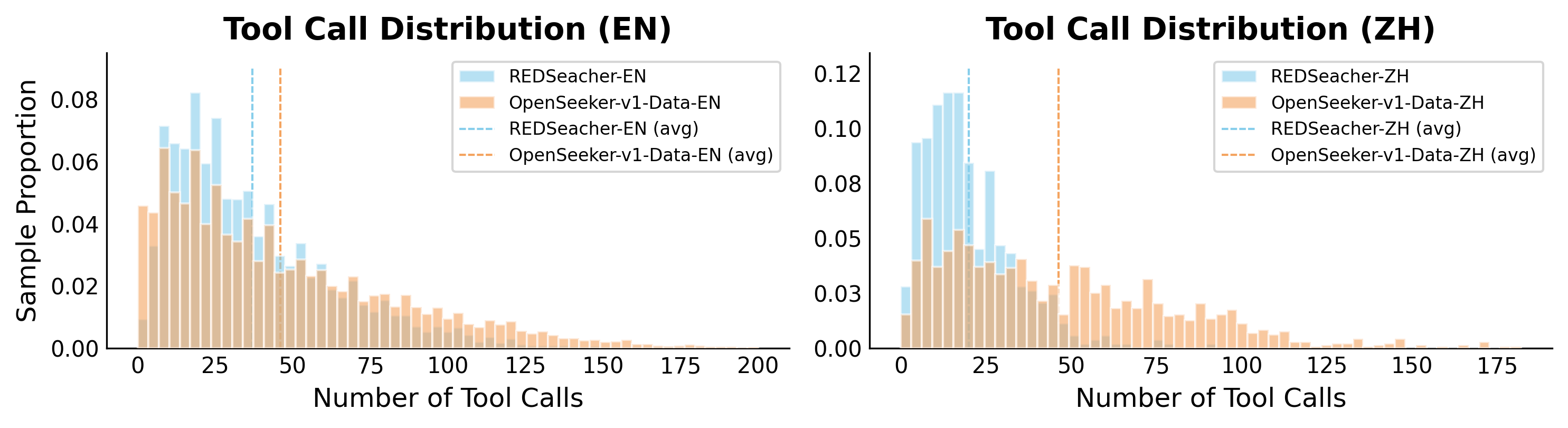}
    \caption{Comparison of tool call counts using the same model for inference: OpenSeeker's data vs. REDSearcher's data. OpenSeeker's data demonstrates a significantly higher average number of tool calls.}
    \label{fig:data_stats_red}
\end{figure}

Quantitative results further validate these advantages. As shown in Table~\ref{tab:model_comp_sft_2}, when trained using the same SFT methodology, OpenSeeker comprehensively outperforms OpenResearcher across three benchmarks. Notably, on BrowseComp-ZH, OpenSeeker surpasses RedResearcher by 21.6\% (48.4\% vs 26.8\%). Furthermore, as illustrated in Figure~\ref{fig:data_stats_red}, when employing the same model for inference to compare tool call counts, OpenSeeker's data proves significantly more challenging than that of RedResearcher. Specifically, the average number of tool calls for OpenSeeker-v1-Data-EN averages 45.92 calls against 36.91 for RedResearcher-EN, while OpenSeeker-v1-Data-ZH is 46.35 compared to 20.02 for RedResearcher-ZH.
\begin{table}[t]
    \centering
    \small 
    \caption{Performance comparison of concurrent works trained via SFT.}
    \label{tab:model_comp_sft_2}
    \scalebox{1}{ 
    \begin{tabular}{l c c c c c c}
        \toprule
        \textbf{Data} & \textbf{\# Samples} & \textbf{\# OS Samples} & \textbf{Academic} & \textbf{BrowseComp} & \textbf{BC-ZH} &\textbf{xbench} \\
        \midrule
        Openresearcher & 96k & 96k & $\times$ & 26.3 & - & 65 \\
        REDSearcher & ? & 10k & $\times$ & \textbf{34.7} & 26.8 & - \\
        \midrule
        \textbf{OpenSeeker-v1-Data-11.7k} & 11.7 k & 11.7k &  \checkmark & 29.5 & \textbf{48.4} &\textbf{74.0}\\
        \bottomrule
    \end{tabular}
    }
\end{table}

\end{document}